\documentclass[conference]{IEEEtran}
\IEEEoverridecommandlockouts

\usepackage{cite}
\usepackage{amsmath,amssymb,amsfonts}
\usepackage{algorithmic}
\usepackage{graphicx}
\usepackage{textcomp}
\usepackage{xcolor}
\usepackage{multirow}
\usepackage{url}
\usepackage{booktabs}
\usepackage{subcaption}
\usepackage{array}
\usepackage{makecell}
\usepackage{stfloats} 
\usepackage{enumitem}
\setlist[itemize]{align=parleft,left=0pt..1em}
\usepackage[belowskip=-4pt,aboveskip=2pt]{caption}

\usepackage[colorlinks=false, urlcolor=black, linkcolor=black]{hyperref}
\def\BibTeX{{\rm B\kern-.05em{\sc i\kern-.025em b}\kern-.08em
    T\kern-.1667em\lower.7ex\hbox{E}\kern-.125emX}}

\newif\ifcomments
\commentsfalse
\commentstrue  

\usepackage{fancyhdr}
\begin{document}

\title{Analyzing Cultural Representations of Emotions in LLMs through Mixed Emotion Survey
}

\author{\IEEEauthorblockN{Shiran Dudy\IEEEauthorrefmark{4},
Ibrahim Said Ahmad\IEEEauthorrefmark{4}, Ryoko Kitajima\IEEEauthorrefmark{2} and
Agata Lapedriza\IEEEauthorrefmark{4}\IEEEauthorrefmark{3}}
\IEEEauthorblockA{
\IEEEauthorrefmark{4}Northeastern University, Boston, MA, USA\\
\IEEEauthorrefmark{2}Independent researcher\\
\IEEEauthorrefmark{3}Universitat Oberta de Catalunya, Barcelona, Spain\\
Email: \{s.dudy,i.ahmad,a.lapedriza\}@northeastern.edu}}


\maketitle
\thispagestyle{fancy}

\begin{abstract}
Large Language Models (LLMs) have gained widespread global adoption, showcasing advanced linguistic capabilities across multiple of languages. There is a growing interest in academia to use these models to simulate and study human behaviors. However, it is crucial to acknowledge that an LLM's proficiency in a specific language might not fully encapsulate the norms and values associated with its culture. Concerns have emerged regarding potential biases towards Anglo-centric cultures and values due to the predominance of Western and US-based training data. This study focuses on analyzing the cultural representations of emotions in LLMs, in the specific case of mixed-emotion situations. Our methodology is based on the studies of Miyamoto et al. (2010), which identified distinctive \textit{emotional indicators} in Japanese and American human responses. We first administer their mixed emotion survey to five different LLMs and analyze their outputs. Second, we experiment with contextual variables to explore variations in responses considering both language and speaker origin. Thirdly, we expand our investigation to encompass additional East Asian and Western European origin languages to gauge their alignment with their respective cultures, anticipating a closer fit. We find that (1) models have limited alignment with the evidence in the literature; (2) written language has greater effect on LLMs' response than information on participants origin; and (3) LLMs responses were found more similar for East Asian languages than Western European languages. 
\end{abstract}


\section{Introduction}
With the widespread adoption of Large Language Models (LLMs), like ChatGPT, there has been an increased interest in understanding how different cultures are represented in LLMs \cite{leong2023bhasa,buttrick2024studying}. Understanding cultural representations in LLMs is crucial for many reasons, such as ensuring that communication tools based on LLMs are inclusive (able to deal with different perspectives and values) and effective (able to interpret and respond appropriately in different cultural contexts). In turn, emotions play a significant role in communication, since they have a strong influence on how messages are expressed and perceived \cite{andersen1996principles}. This has motivated recent works on understanding emotional skills in LLMs \cite{schaaff2023exploring,tran2023robustness,ghosal-etal-2020-cosmic,broekens2023fine}. 

This paper explicitly studies \textit{cultural representations of emotions} in LLMs. Concretely, we focus on comparing emotion representations of Western and East Asian cultures, by studying the emotional responses of LLMs to various mixed-emotion situations. Our research builds upon the \emph{Mixed Emotion Experiment} conducted by Miyamoto et al. (2010)~\cite{miyamoto2010culture}. In their work, the authors designed a survey describing 13 situations of mixed emotions, where the participants had to rate how they would feel in each of the situations. The survey was answered by North American participants and Japanese participants, in their respective languages, and the results show interesting differences between the two studied populations (more details on the \emph{Mixed Emotion Experiment} \cite{miyamoto2010culture} are provided in  \ref{sec:related_work_mixed}). In our work, we utilize the same emotional response survey as in \cite{miyamoto2010culture} in both Japanese and English languages to investigate whether the responses generated by various LLMs align with the findings observed in the human experiments conducted by Miyamoto et al. (2010)~\cite{miyamoto2010culture}. Our work is closely related to the work by Havaldar et al. (2023)~\cite{havaldar2023multilingual}, who investigated culturally aware emotional responses in LLMs. This work investigates the mixed emotion phenomenon, rather than the presence of universal emotions in LLMs.

More generally, our study addresses three research questions:

\begin{itemize}
\item \textbf{RQ1}: To what extent the findings in~\cite{miyamoto2010culture} are reproduced with LLMs?
\item \textbf{RQ2}: What is the effect on LLMs' response when prompted with different sources of contextual information?
\item \textbf{RQ3}: How similar are the responses of LLMs in languages with higher cultural affinity?
\end{itemize}

To address these research questions we perform 3 studies, which are presented in Sect. \ref{sec:experiements}. \textit{Study 1} compares how the responses to the survey differ when the LLMs are prompted in English vs. Japanese. It also studies to what extent the results obtained by LLMs align with the results obtained by Miyamoto et al. \cite{miyamoto2010culture} on their experiments with human subjects. \textit{Study 2} analyzes the effect of adding different contextual sources to the prompt. For example, prompting in English language but adding to the prompt the textual information \emph{Please rate [the survey] as a Japanese participant}. Finally, \textit{study 3} compares the responses across different languages, including 4 Western languages (English, Spanish, German, and French) and 4 East Asian languages (Japanese, Chinese, Korean, and Vietnamese).

Our studies are conducted on 5 widely used LLMs: three open source models 
(mistral-7b-instruct~\cite{jiang2023mistral}, gemma-7b-it:free~\cite{team2024gemma}, and llama-2-70b-chat~\cite{touvron2023llama}) and two of the most popular private systems (gpt-3.5-turbo\footnote{\url{https://platform.openai.com/docs/models}}
and gpt-4-turbo-preview\footnote{\url{https://openai.com/contributions/gpt-4v}}). Our results show that LLMs have limited alignment with our investigated phenomenon (i.e. model's responses to the survey do not show a strong alignment with the results obtained with human experiments). We also found that language has a stronger effect on responses generated by LLMs, compared to context about the user's origin, and that LLM responses for East Asian languages were found to correlate with each other more than the European ones. Besides these interesting observations, we think that the methodology used in this paper could inspire future studies on cultural representations in LLMs.

\section{Related Work}
\label{sec:related_work}

\subsection{Evaluating emotional skills of Large Language Models}

Some recent studies have attempted to evaluate and quantify emotional skills in Large Language Models (LLMs). For example, Schaaff et al. 2023 \cite{schaaff2023exploring} investigates the extent to which ChatGPT based on GPT-3.5 can exhibit empathetic responses and emotional expressions. In the experimentation, ChatGPT was instructed to rephrase neutral sentences into six emotional sentences of joy, anger, fear, love, sadness, and surprise. The responses of ChatGPT were then assessed using five standardized questionnaires. This study demonstrated promising results in ChatGPT's ability to convey multiple emotions using the same base sentence. 
 

Tran et al. \cite{tran2023robustness} introduce a method to assess robustness and bias in emotion recognition systems. They combine GPT-3 and rule-based constraints to create text variations for different language proficiency levels. These texts are used to examine performance biases, focusing on language proficiency, and investigate model-agnostic effects using the COSMIC model \cite{ghosal-etal-2020-cosmic} and benchmark datasets.

Finally, Broekens et al. \cite{broekens2023fine} explore how ChatGPT can perform affective computing tasks using prompting alone. The methodology involves conducting conversational experiments with ChatGPT to explore its fine-grained affective processing capabilities. The paper used a rule-based logical model of appraisal as a prompt based on the OCC model~\cite{ortony2022cognitive} to assess if ChatGPT can predict emotions according to a specific framework. They also mapped stimulus sets using human expert raters as ground truth and created a set of situations reflecting different emotions in the OCC model. The study revealed that ChatGPT can accurately extract fine-grained sentiment from situations and words, showing comparable performance to fine-tuned models. It demonstrated a moderate understanding of affective dimensions and emotion words, and successfully performed basic emotion elicitation based on the OCC appraisal model. 

\subsection{Cross-cultural emotion studies}

Emotions were extensively studied comparing Eastern Asian culture to Western culture. Tsai et al. \cite{tsai2006cultural} showed that people in North American contexts lean towards feeling excited, enthusiastic, energetic, and other “high arousal positive” states, while people in East Asian contexts, generally prefer feeling calm, peaceful, and other “low arousal positive” states. A follow-up by Tsai et al. \cite{tsai2007influence} showed that people from North American contexts (who value high arousal affective states) tend to prefer thrilling activities like skydiving, whereas people from East Asian contexts (who value low arousal affective states) prefer tranquil activities like lounging on the beach.

In addition, Tsai et al. \cite{tsai2006cultural} showed that among European Americans, the less people experience high arousal positive states, the more depressed they are, while, among Hong Kong Chinese, the less people experience low arousal positive states, the more depressed they are. In line with the previous insights, Chentsova-Dutton et al. \cite{chentsova2007depression} found that depressed European Americans show reduced emotional expressions, but depressed East Asian Americans do not and, in fact, may express more emotion. In our work, we focus on Miyamoto et al. \cite{miyamoto2010culture}, who compared North Americans to Japanese and showed the latter group is more likely to feel bad and good (“mixed” emotions) during positive events and differ in response when facing negative events.  

\subsection{Cultural representations in LLMs}
\label{sec:related_work-cultural}

A few previous studies have already analyzed cultural representations in LLMs. For example, Naous et al. \cite{naous2023having} assessed biases in LLMs towards Arab and Western cultures, focusing on story generation, NER (named entity recognition), sentiment analysis, contextual prompt analysis, and text infilling. The study shows that multilingual and Arabic monolingual LMs exhibit bias towards entities associated with Western culture. The study further introduced a resource called CAMeL containing naturally occurring prompts and entities that contrast Arab and Western cultures. In line with that, Atari et al. \cite{atari2023humans} show that the LLMs performance of psychological cognitive tasks resembles most of the Western Industrialized Educated Rich and Democratic (WIERD) population in an extensive cross-language study on LLMs. Similarly, Arora et al. \cite{arora2023probing} found in their study which was based on World Values Survey\footnote{\url{https://www.worldvaluessurvey.org/WVSContents.jsp}}, that while the values elicited from models vary across cultures, their bias is not in line with values outlined in existing large-scale values surveys. More widely, LLMs cultural representation has started gaining interest in mainstream media outlets \cite{Piir2023Finland,Akira2023i}.

Beyond output analysis, a recent study from Wendler et al. \cite{wendler2024llamas} explored whether multilingual language models use English as an internal pivot language, focusing on the Llama-2 family of transformer models, probing LLMs internally. The study was done by tracking intermediate embedding and logit lens analysis based on constructed non-English prompts with a unique correct single-token continuation. The study provides compelling evidence that multilingual language models may indeed use English as an internal pivot language, albeit in a nuanced and conceptually biased manner.

\section{Methods}

\subsection{The Mixed Emotion Experiment with Human Participants
}
\label{sec:related_work_mixed}

In this work we follow the protocol set by Miyamoto et al. (2010) \cite{miyamoto2010culture} who studied the similarities and differences in which Americans and Japanese experience mixed emotions, focusing on the co-occurrence of positive and negative emotions experienced by Americans and Japanese in plausible situations of their daily lives. Concretely, the study considers 13 situations, which are divided in three types: self-success, transition, and self-failure. Then, participants need to rate the extent to which they would experience positive or negative emotions, and the extent they would experience specific positive emotions (happiness, pride, sympathy, relief, hope, and friendly feeling) or specific negative emotions (sadness, anxiety, anger, self-blame, fear, anger at oneself, shame, guilt, jealousy, frustration, embarrassment, resentment, and fear of troubling someone else). Additionally, participants answered three appraisal questions using 6-point scales: (1) how responsible they would feel for other people’s feelings; (2) how much other people were responsible for their feelings. Finally, to study the participants’ motivation to control the situation, participants were asked \emph{how much would you think about influencing or changing the surrounding people, events, or objects according to your own wishes?} using also a 6-point scale. The obtained results showed that mixed emotions were more present for the Japanese participants than for the Americans in the self-success situations. Beyond mixed emotions, additional significant differences were found between those populations in self-failure situations when compared on the motivation to change the situation, and the self-responsibility attributed which will be discussed in the next sections.  

Our work follows the study protocol used in \cite{miyamoto2010culture}. Concretely, we replicated the survey focusing on questions that demonstrated significant difference between Japanese and Americans in that study.
We contrast our findings with findings in \cite{miyamoto2010culture} which are summarized in Table~\ref{tab:emotional_responses}.

\subsection{Running the Mixed Emotion Survey on LLMs} 

Based on the original examples in \cite{miyamoto2010culture} we generated five self-success situations and five self-failure situations with ChatGPT, which composing the survey, while incorporating the original instructions, as described in the paper. The following two situations are examples of Self-Success and an example of Self-Failure, respectively: 

\begin{itemize}
\item(Self-Success situation) \emph{You receive a stellar performance review and a promotion, which makes you happy. However, your colleague receives a warning due to underperformance, leaving you with mixed emotions.}
\item (Self-Failure situation) \emph{Your close friend becomes the center of attention at social gatherings, effortlessly making friends and connections, which fills you with pride for their social skills. Meanwhile, you struggle to navigate social situations and feel left out, resulting in mixed feelings of admiration for them and disappointment in yourself.}
\end{itemize}
 
For \textit{study 3}, each survey was 
translated into Vietnamese, Korean, Chinese, French, German, and Spanish, by native speakers who also hold proficiency in English. To query the LLMs, we employed LangChain python package~\footnote{\url{https://python.langchain.com/docs/get_started/introduction}} and only included complete responses for all 24 questions. The survey questions were split into three parts, as currently, models fail to answer all 24 at once. We evaluate the emotional response for a situation assuming each can be asked separately per situation, and that the presence of other questions does not influence the response. 

\subsection{Evaluations}
We employed independent paired one-tailed in \textit{study 1}, two-tailed t-tests in \textit{study 2} and \textit{3}, when comparing distributions. Regarding \textit{study 1} and, particularly, the comparison of LLMs with the human experiments performed by Miyamoto et. al \cite{miyamoto2010culture}, we compare our results with the relevant findings of \cite{miyamoto2010culture}, which are summarized in Table.~\ref{tab:emotional_responses}.

\begin{table}[htbp]
\centering
\caption{Results of the experiments with human participants obtained by Miyamoto et al. (2010) \cite{miyamoto2010culture}. The main findings were on two situation types, covering different emotions. Their findings are based on a one-tailed t-test (with their associated $p$ values) indicative of the relation between groups (JP or AM that correspond to Japanese and Americans).}
\label{tab:emotional_responses}
\begin{tabular}{@{}llll@{}}
\toprule
\multicolumn{1}{c}{\textbf{Situation Type}} & \textbf{Emotion} & \textbf{Relation} & \textbf{p-value}  \\ \midrule
\multirow{5}{*}{Self-Success} & Motivation to change & JP $>$ AM &  $p<0.05$  \\
& Me responsible for others & JP $>$ AM &  $p<0.01$  \\

& Happy and  &   & \\
& Fear of troubling others & JP $>$ AM &  $p<0.01$  \\ 
& Positive and negative & JP $>$ AM &  $p<0.1$\\ \midrule
\multirow{3}{*}{Self-Failure} & Motivation to change & JP $<$ AM &  $p<0.001$  \\
& Me responsible for others & JP $>$ AM &  $p<0.001$  \\
& Others Responsible for me & JP $<$ AM &  $p<0.07$ \\ \bottomrule
\end{tabular}
\end{table}

\section{Experiments}
\label{sec:experiements}

\subsection{Study 1: English vs. Japanese}
We evaluate the differences in the survey responses of the LLMs when prompted in English vs. Japanese. We also compared the responses of LLMs with the results obtained by Miyamoto et al. \cite{miyamoto2010culture} in their study with human subjects. 

We run the survey $n$ times per language, per LLM, to simulate $n$ responses. We searched for $n$ that is stable such that, the next time we draw $n$ responses, the distribution of both drawings is similar -- indicating sample stability. We found $n=80$ to provide stable distributions and elaborate on the search procedure in section~\ref{sec:findingn}.

Table~\ref{tab:emotional_responses_llm} summarizes the results per system for each of the emotion responses presented in Table~\ref{tab:emotional_responses}. 
To produce the table, we conducted one-tailed t-tests with \(H_0\) hypothesis, positing no difference between the groups, and \(H_1\) indicating a difference in the direction specified in Table~\ref{tab:emotional_responses}. If the t-test outcome rejected \(H_0\) in the expected direction, we denoted it with a `+' sign. Subsequently, we conducted another t-test in the opposite direction, and if the outcome rejected \(H_0\), it was indicated with a `-' sign. If \(H_0\) was not rejected, we left the cell empty. Following the receipt of results, we aggregated the overall performance, with `+' and `-' counting for \(+1\) and \(-1\) respectively across the 7 tests. Finally, the results were normalized to a percentage score within the range \([-100\%,100\%]\), corresponding to total failure and total success across all tests.

\begin{table*}[htbp]
\centering
\caption{Emotional Responses across five state-of-the-art LLMs. Each LLM is marked with `+', `$-$', or an empty cell that corresponds to the relation found between Japanese to Americans. For instance, in gemma for motivation to change in self-success situations, one-tailed t-test was significant, and \textit{in the direction of the original study} shown in Table~\ref{tab:emotional_responses} (JP$>$AM), resulting in `+'. On the other hand, for motivation to change in self-failure situations,  gemma showed a significant result \textit{in the opposite direction} of the original study (gemma: JP$>$AM, Table~\ref{tab:emotional_responses}: JP$<$AM), and therefore is marked with `-'. For non-significant results in either direction, the cell is empty. At the bottom, we aggregate the number of `+', and `$-$'. Finally, reading the summary for example can be the following for gpt4. gpt4 was in-line with the original study outcome 3 times, but was in the opposite 3 times, resulting in $0$. Results were normalized.}
\label{tab:emotional_responses_llm}
\begin{tabular}{|ll|c|c|c|c|c|c|}
\hline
\textbf{\makecell{Situation\\Type}} & \textbf{Emotion} & \textbf{\makecell{mistral\\-7b-Instruct}} & \textbf{\makecell{gemma\\-7b-IT:Free}} & \textbf{\makecell{llama\\-2-70b-Chat}} & \textbf{\makecell{gpt-3.5\\-Turbo}} & \textbf{\makecell{gpt-4\\-Turbo-Preview}} & \textbf{Original Study} \\ \hline
\multirow{5}{*}{\makecell{Self-\\Success}} & Motivation to change & + & + & + &  + & $-$ & $p<0.05$\\ \cline{3-8} 
& Me responsible for others & + & $-$ & $-$ & + & + & $p<0.01$\\ \cline{3-8} 
& Happy and   & & & & & &  \\ 
&  Fear of troubling others & + & + & + & + & $-$ & $p<0.01$\\ \cline{3-8}
& Positive and negative & $-$ & & & & & $p<0.1$\\  \hline
\multirow{3}{*}{\makecell{Self-\\Failure}} & Motivation to change & & $-$ & $-$ & $-$ & + & $p<0.001$  \\ \cline{3-8} 
& Me responsible for others & $-$ & $-$ & $-$ & + & +  & $p<0.001$\\ \cline{3-8} 
& Others responsible for me & $-$ & + & + & $-$ & $-$ & $p<0.07$\\ \hline
\multicolumn{1}{c}{\vspace{0.01cm}}\\ \hline
 &  \makecell{Performance count\\
 $x\in[-7,7]$} & $3-3=0$ & $3-3=0$ & $3-3=0$ & $4-2=2$ & $3-3=0$  & \\  \hline
  &  \makecell{Normalized success \\$x\in[-100\%,100\%]$} & $0\%$ & $0\%$ & $0\%$ & $28.5\%$ & $0\%$  & \\  \hline
\end{tabular}
\end{table*}

\begin{figure*}
  \centering
  \includegraphics[width=1\textwidth]{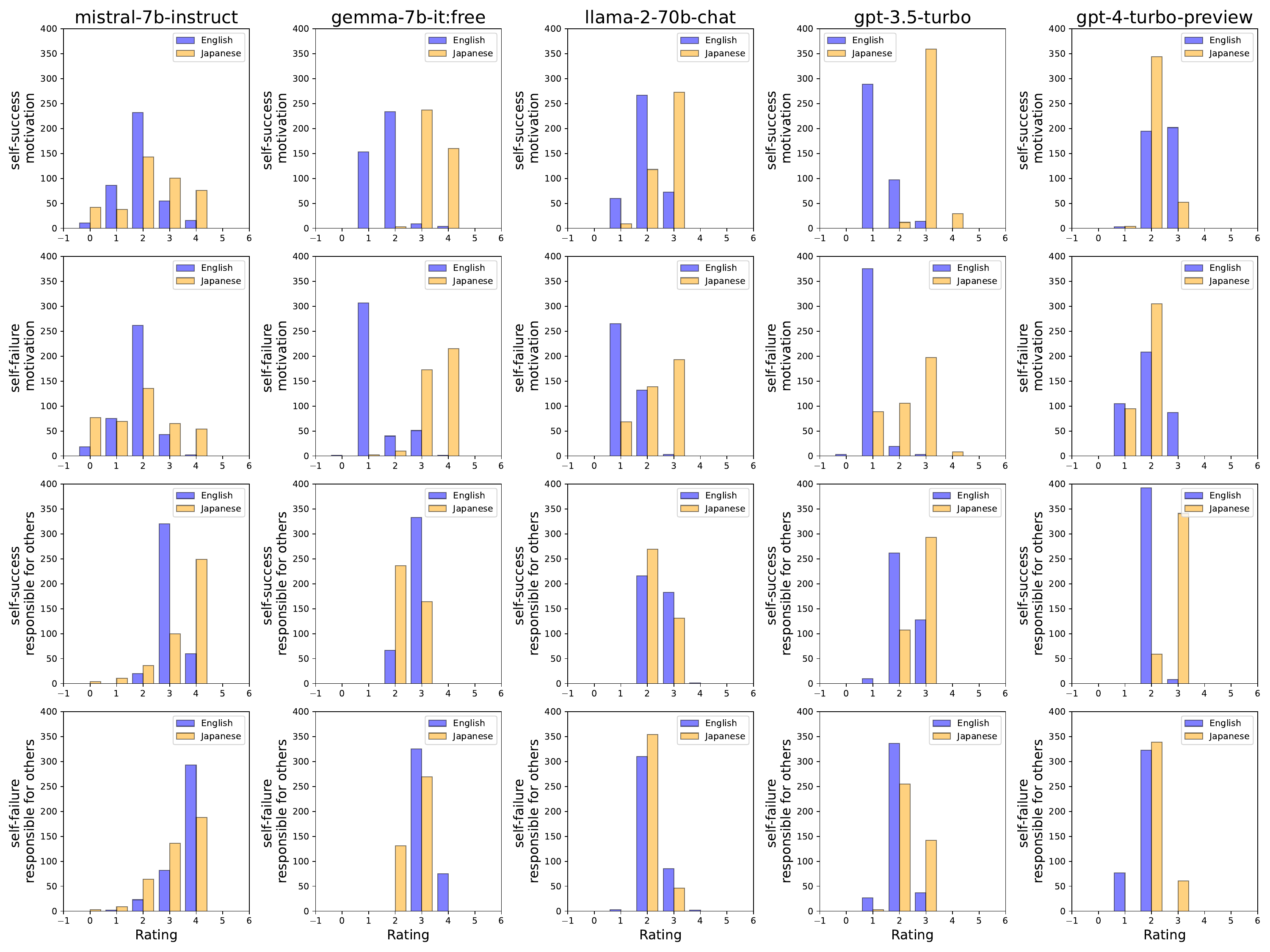}
  \caption{Emotion responses by LLM. This figure presents two different emotions (`motivation to change', and `me responsible for others' under two different situations (self-success and self-failure) across five LLMs. First, per row, we can notice that LLMs underlying distributions is different. gemma, in the first row, for example, offers a clear separation for Japanese and Americans, where in mistral they are mixed. These differences may indicate that the different LLMs may not have the same underlying mechanism. In addition, we can also see that gemma, llama, and gpt3.5 may not be just translating prompts from English due to the relative differences in the two populations.}
  \label{fig:raw_dist}
\end{figure*}

In Table~\ref{tab:emotional_responses_llm}, we observe that most models achieved a similar balance of the number of tests aligned with the human subject study (\ref{tab:emotional_responses}). GPT-3.5 attained \(28.5\%\) indicating greatest alignment compared to other models, yet it did not demonstrate full alignment with the results from the original literature. To further investigate the responses of the LLMs, Figure~\ref{fig:raw_dist} describes four types of rating distributions per LLM: \textit{(self-success, motivation to change)}, \textit{(self-failure, motivation to change)}, \textit{(self-success, me responsible for others)}, and \textit{(self-failure, me responsible for others)}. We notice differences in distributions per \textit{(self-success, motivation to change)}, which is in the first row. Gemma indicates a clear seperation between Japanese and Americans, where mistral's populations is more mixed. We see that even though most models align with human experiments on this survey question in Table~\ref{tab:emotional_responses_llm}, the distributions seem visually different, with a greater or smaller degree of separation of the two populations in gemma, llama, gpt3.5, and gpt4. This may suggest that their underlying mechanism may be different. The fact we find separation may also suggest that even if LLMs do translate under-the-hood, they may not just do so and exhibit a level of cultural sensitivity to this question. 

Conversely, the first two rows in Figure~\ref{fig:raw_dist} exhibit visually similar distributions, even though they correspond to self-success and self-failure scenarios, where one would expect opposite responses as indicated in Table~\ref{tab:emotional_responses}. This similarity may suggest a reduced sensitivity to the \textit{situation type}.


The last two rows in Figure~\ref{fig:raw_dist} also demonstrate similar distributions per model for different situations types. In this case though, Table~\ref{tab:emotional_responses} indicates that the relation between Japanese and Americans remained the same. Therefore, these responses are aligned with the findings.

In light of these two findings we cannot discern whether there is insensitivity to situation type, or this was a one-off mistake by all models.

Overall, we found a limited alignment of LLM responses with Miyamoto et al. \cite{miyamoto2010culture}. This limited alignment has been exhibited variably across models, manifesting in different responses to various tests. However, similarities were also observed, with many models showing low sensitivity to the same textual information.


\textit{Finding the number $n$ of required samples:}\label{sec:findingn} While Miyamoto et al. \cite{miyamoto2010culture} emphasize the importance of the number of human participants to obtain relevant results, our study focuses on ensuring the stability of responses from LLMs to enable reproducibility (on that version).
We accomplish this by executing the same experiment $n$ times and then computing our results based on the average of the responses obtained along the $n$ runs. The last part of this sub-section details the experiments we performed to determine the parameter $n$ that we use in all the experiments. Concretely, we determine the minimal number \( n \) of samples drawn from a model to ensure sample stability, meaning that the distributional properties ($\mu$, $\sigma$) remain consistent when drawing a new set of $n$ samples.

\begin{figure}
  \centering
  \includegraphics[width=0.5\textwidth]{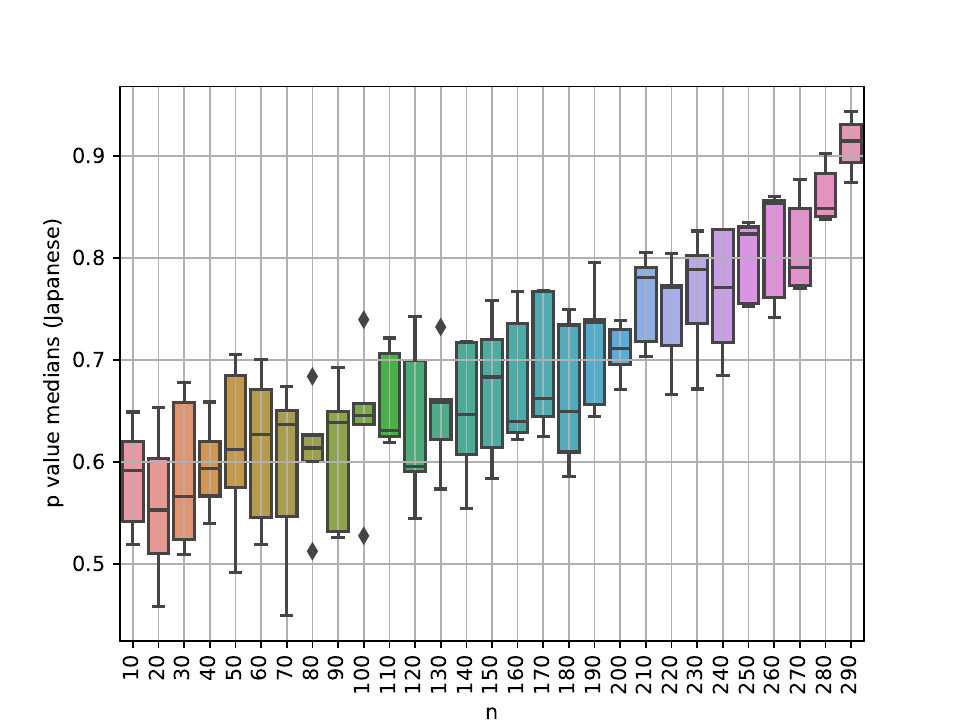}
  \caption{Experiments to determine the number of necessary runs ($n$) to ensure stability of the LLMs' output across the different models. We define stability when drawing twice (or more) $n$ responses, a paired t-test does not show a low \textit{p-value}. This $n$ was searched separately for Japanese and English responses, and across the LLMs we experimented with. Empirically, we searched for $n$ that its \textit{p-value} median distribution (shown in each boxplot) is above $0.5$, and this figure shows that for Japanese, which was less stable than English, $n=80$ to achieve stability. Therefore $n=80$ was fixed throughout all our experiments.}
  \label{fig:jp_n}
\end{figure}

To determine \( n \), we administered a subset of the survey questions at intervals of 10, ranging from 10 to 300 samples. For each \( n \), we generated 20 distributions and conducted two-tailed t-tests for all unique pairs of distributions. Figure~\ref{fig:jp_n} presents boxplots for Japanese, showing the \( p \)-value medians across all models for each \( n \) (five $p$-value medians per $n$ that each corresponds to a different model). Empirically, we observed that $n$s resulting in median \( p \)-values above 0.5 provided stable distributions. Based on that observation the minimum \( n \) required was determined to be 80. 

\subsection{Study 2: English vs. Japanese using context prompts.}

The goal of this part is to understand the effect of different types of contexts on the response. We set to investigate the effect of two modes of context: 1) the language of the \textit{written} survey indicated in ($w$), and 2) the language of \textit{origin} of the speaker (which the LLM is required to simulate), indicated in ($o$) forming a response distribution described as $(w,o)$. 
We added to the survey an additional request `Please rate it as a [placeholder] participant', where the placeholder could be either `Japanese' or `American' which were applied to both languages -- totaling in four combinations of written language and origin $(w,o)$: $(en,en)$,$(en,jp)$,$(jp,jp)$,$(jp,en)$. For instance $(en,jp)$ is a distribution made by a survey written in English simulating a participant from Japan. We hypothesized that due to either similarity in language or similarity in the origin of the simulated participant, we might find correlations across different ($w$,$o$) combinations. For every two pairs of ($w$,$o$) we ran two-tailed t-tests to evaluate distribution similarity across the 7 t-test shown in Table~\ref{tab:emotional_responses}. For instance, in Table~\ref{tab:results_conditions} the responses by gemma model comparing ($(en,jp)$) and $(en,en)$) were similar in 3 of the 7 tests applied. Following this table, we posed three hypotheses to understand the effect of the written language ($w$), and participant's origin ($o$) that are shown in Table~\ref{tab:hypotheses_conditions}: (1) are there similarities between shared language and different origin?; (2) are there similarities between similar origin, but different language? and; (3) are there similarities in responses between the response in \textit{study 1} ($w$,-), where origin was not indicated, to responses that shared that same written language ($w$,$o$) in this section?

Table~\ref{tab:hypotheses_conditions} summarizes the findings based on the raw results in Table~\ref{tab:results_conditions}.We find that question (1) elicited the greatest number of correlations across most models, indicating a strong effect that can be attributed to the written language compared to the shared origin (2) as contextual source. However, it was surprising the to learn that hypothesis (3) was not true, as we did not find any correlation between \textit{study 1} and the same language in \textit{study 2}. 

Mistral presented the highest number of similarities, yet still half of the expected correlations were missing. In  Table~\ref{tab:results_conditions} we find that mistral also demonstrated similarities between $(en,en)$ and $(jp,jp)$ which actually may be a warning sign for possibly generating too correlative distributions overall. Therefore, despite Mistral's high rate of correlation, it may suggest greater similarity across responses, which is contrary to cultural sensitivity. 

In this study we find that language has a stronger effect on responses, than the participant's origin (or possibly other textual information). Yet, when a shared language was compared with \textit{study 1} responses, we didn't find any correlations, which make these responses inconsistent in behavior.

\begin{table}
\centering
\caption{Textual language crossed with participant origin. This table is a confusion matrix, where each ($w$=written language,$o$=participant's origin) indicates a single response to a survey. For instance, \textit{(en,jp)} is an LLM response to a survey written in English, and the participant's origin is from Japan. Since we compared the results to \textit{study 1}, where we didn't indicate any origin the notation was \textit{(jp,-)} referring to the Japanese survey in \textit{study 1}. Each of ($w$, $o$) distribution was compared with another one in the table. For instance, \textit{(en,en)} was correlating with \textit{(jp,en)} only in mistral, where 2 of the 7 tests were found similar.}
\label{tab:results_conditions}
\begin{tabular}{|c|c|c|c|c|c|}
\hline
($w$, $o$)&  (en, en) & (en, jp) & (jp, en) & (jp, jp) & (jp, -)\\ \hline
(en, -)  & & & & & \\ \cline{1-6}
(en, en)  & & \makecell{mistral: 7\\ gemma: 3 \\ llama: 6 \\ gpt3.5: 4} & mistral: 2 & mistral: 3 &\\ \cline{1-1} \cline{3-6}
(en, jp)  & & & \makecell{mistral: 2\\gpt4: 1} & mistral: 3 &\\ \cline{1-1} \cline{4-6}
(jp, en)  & & & & \makecell{mistral: 7\\ gemma: 5\\ llama: 3\\gpt3.5: 2\\gpt4: 2} &\\ \cline{1-1} \cline{5-6}
(jp, jp)  & & & & & \\ \cline{1-1} \hline
\end{tabular}
\end{table}

\begin{table*}[htbp]
\centering
\caption{Evaluating the effect of different contexts. Post-processing Table~\ref{tab:results_conditions}. Here we evaluated how similar distributions are due to shared written language ($w_0=w_1$ shown in a.) same origin of participant ($o_0=o1$ shown in b.), or similarity to the experiment in \textit{study 1} (c.), with same language in \textit{study 2}. For example, for hypothesis a. gpt4 had only one correlation found which was between (jp,jp) and (jp,en).}
\label{tab:hypotheses_conditions}
\begin{tabular}{|l|c|c|c|c|c|}
\hline
 \textbf{$H_0$} & \textbf{Mistral-7b-Instruct} & \textbf{Gemma-7b-IT:Free} & \textbf{Llama-2-70b-Chat} & \textbf{GPT-3.5-Turbo} & \textbf{GPT-4-Turbo-Preview} \\ \hline
a. \(w_0 = w_1, o_0 \neq o_1\) & \makecell{1.(\textbf{en},en) and (\textbf{en},jp)\\2.(\textbf{jp},jp) and (\textbf{jp},en)} & \makecell{1.(\textbf{en},en) and (\textbf{en},jp)\\2.(\textbf{jp},jp) and (\textbf{jp},en)} & \makecell{1.(\textbf{en},en) and (\textbf{en},jp)\\2.(\textbf{jp},jp) and (\textbf{jp},en)} & \makecell{1.(\textbf{en},en) and (\textbf{en},jp)\\2.(\textbf{jp},jp) and (\textbf{jp},en)} & 1.(\textbf{jp},jp) and (\textbf{jp},en)\\ \hline
b. \(w_0 \neq w_1, o_0 = o_1\) &  \makecell{1.(en,\textbf{jp}) and (jp,\textbf{jp})\\2.(en,\textbf{en}) and (jp,\textbf{en})} & & & & \\ \hline
c. \(w_0 = w_1, o_0=\emptyset, o_1 \) &  & & & & \\ \hline
\multicolumn{1}{c}{\vspace{0.01cm}}\\ \hline
\makecell{Performance summary\\ $x\in[0,8]$} & 2+2=4 & 2 & 2 & 2 & 1\\ \hline
 \makecell{Normalized success \\$x\in[0\%,100\%]$} & $50\%$ & $25\%$ & $25\%$ & $25\%$ & $12.5\%$\\ \hline
\end{tabular}
\end{table*}

\subsection{Study 3: Comparing East Asian vs. Western Languages}

Schimmack et al. \cite{schimmack2002cultural} claimed that `it is possible that people in Asian dialectic cultures more readily recognize the pleasant and unpleasant aspects of an event, which produces mixed feelings of pleasant and unpleasant emotions.' This hypothesis has to do with the collectivist (vs. individualistic) nature of East Asian cultures \cite{Charlotte2023Understanding,grossmann2017mixed}. In \textit{study 3}, we analyze the similarities across languages of East Asian and American-European cultural affinities anticipating closer similarity with Japanese and American cultures, respectively.

We expanded our survey to include three additional East Asian languages (Chinese (ch), Korean (kr), and Vietnamese (vt)), and three European languages (French (fr), German (gr), and Spanish (sp)). We translated the survey in \textit{study 1}, and compared pairs of languages under the hypothesis that LLM responses to related languages would be similar. Similarly to \textit{study 2}, we conducted 7 t-tests across all language pairs. We summarized our results in Table~\ref{tab:hypotheses_languages}.

We observe that most LLMs generated more similar response distributions for East Asian languages than for European languages. On the one hand, this may be a desired outcome, as it may reflect LLMs may be culturally aware of these languages. On the other hand, based on the Internet Society Foundation data~\footnote{\href{https://tinyurl.com/5yuczd4c}{Internet data distribution by language}}, the most used languages on the internet are English ($55\%$), Spanish ($5\%$), German, French, and then Japanese and Chinese, which may offer a proxy for training data available for LLMs, that possibly resulted in less fine-grained responses to differentiate one East Asian culture from another. Since we see less correlation for European languages, we hypothesize that the larger amount of data help to better distinguish European languages. 

Similar to \textit{study 2}, no language correlated with English, may be due to its disproportional volume ($55\%$) reflecting mostly American culture, and its many facets, in training.\footnote{see training datasets in ~\cite{brown2020language} and \href{https://www.statista.com/statistics/1279537/digital-content-creators-worldwide-by-country/}{number of content creators}}

Here, similar to \textit{study 2}, mistral demonstrated the largest number of similar pairs, resulting in highest number of cross-cultural correlations. 
Based on \textit{study 2} and \textit{study 3}, we suggest that Mistral exhibits a pattern of excessive similarity across responses, regardless of language and textual variation. This indicates that Mistral's responses show low sensitivity to cultural, content, and language differences.


\begin{table*}[htbp]
\centering
\caption{Evaluating similarity related languages. This table summarizes the results across languages for the five LLMs. We hypothesized that related East Asian languages may be found similar based on two-tailed t-tests (shown in a.), and hypothesized similarity for European languages (shown in b.). We shortened languages names to vt, kr, ch, jp, fr, gr, sp, en, corresponds to Vietnamese, Korean, Chinese, Japanese, French, German, Spanish, and English. For example, we found that mistral in hypothesis b. had two correlations: one German and French (gr,fr), and another between German and Spanish (gr,sp) resulting in two. Overall there were more similarities for the East Asian responses than European languages. Since East Asian language resources are smaller on the internet, there is a concern about how well LLMs differentiate within those languages and how well they collapse cultural identity to East Asian}
\label{tab:hypotheses_languages}
\begin{tabular}{|l|c|c|c|c|c|}
\hline
 \textbf{$H_0$} & \textbf{Mistral-7b-Instruct} & \textbf{Gemma-7b-IT:Free} & \textbf{Llama-2-70b-Chat} & \textbf{GPT-3.5-Turbo} & \textbf{GPT-4-Turbo-Preview} \\ \hline
a. \(w\in W_{East Asian}\)    & 1.(kr,vt), 2.(ch,jp), 3.(jp,vt) & 1.(kr,vt), 2.(jp,vt), 3.(ch,vt) & 1.(kr,vt) & 1.(jp,kr) &  1.(jp,kr), 2.(jp,vt)\\ \hline
b. \(w\in W_{European}\) & 1.(gr,fr), 2.(gr,sp) & 1.(gr,fr), 2.(gr,sp) &  & 1.(gr,fr) & \\ \hline
\multicolumn{1}{c}{\vspace{0.01cm}}\\ \hline
\makecell{Performance summary\\ $x\in[0,12]$} & 3+2=5 & 3+2=5 & 1 & 1+1=2 & 2 \\ \hline
 \makecell{Normalized success \\$x\in[0\%,100\%]$} & $41.6\%$ & $41.6\%$ & $8.3\%$ & $16.6\%$ & $16.6\%$\\ \hline
\end{tabular}
\end{table*}


\section{Conclusions}

In this work we study the cultural alignment of Large Language Models (LLMs) in the context of mixed emotions. The mixed emotions phenomenon is linked to Eastern collectivist norms \cite{miyamoto2010culture}. In the study conducted by Tim Lomas et al. \cite{lomas2023complexifying}, the central premise is that individualism is typically associated with Western cultures, whereas collectivism is more closely linked to Eastern cultures. Building on this premise, we aimed to investigate the sensitivity of LLMs to cultures beyond Western contexts. We did this by examining LLM responses to mixed emotion situations, which are believed to elicit either collectivist or individualistic norms depending on the cultural proximity to the originally studied cultures.

Our findings indicate that, when replicating the human participants' study from Miyamoto et al. \cite{miyamoto2010culture} in English and Japanese, the responses of leading LLMs demonstrated limited alignment to the humans' responses obtained in \cite{miyamoto2010culture}. 

We also compared the effect of prompting the LLMs in English or Japanese versus explicitly using textual descriptions of the cultural context. The goal of these experiments was to get answers from the model as if the questions were posed to individuals from Western cultures versus Japanese culture. We discovered that the language itself was impacting the responses more than its textual context description. 

Finally, we evaluated the responses of the LLMs when prompted in various languages, including East Asian languages  (Chinese, Korean, and Vietnamese) as well as American European (or Western) languages (French, German, and Spanish). We found that the similarity of responses across the investigated LLMs varied significantly between East Asian and Western languages. The East Asian language cohort exhibited a greater number of correlations in responses, making it more similar compared to the Western language cohort. This was inconsistent with our expectation that both cohorts would show similar rates of correlation.

As LLMs become more widely accessible globally, and as more researchers explore their potential to simulate human behavior, there is a growing need to understand how accurately they reflect our diverse values and cultures.
We hope that the methodology presented here, which replicates peer-reviewed human-subject studies, offers a path for advancing our understanding of cultural alignment based on the extensive literature on various cultures the research community has accumulated. Furthermore, we hope that the observations and discussions of our findings will inspire more research into understanding cultural representations of emotions in LLMs.

\section*{Ethical Impact Statement}
\textbf{Limits of Generalizability (applicability to other cultures)}.
Our approach relied on existing peer-reviewed literature, limiting our exploration of less well-researched cultures. For these under-researched cultures, we recommend developing surveys that focus on understanding cultural norms and emotional differences (with respect to dominant cultures trained by LLMs) to more effectively evaluate the cultural alignment of human subject responses to LLMs.

\textbf{Limits of Generalizability (of broader cultural representation)}.
While the findings were significant according to standard statistical methods, we evaluated the mixed emotions phenomenon based on seven tests. These results highlight specific cultural differences related to mixed emotions but do not allow for broader conclusions about overall cultural representation.

\textbf{Validation}. One concern is the validity of the experiment conducted by Miyamoto et al. \cite{miyamoto2010culture} and we recognize there might be shifts in cultural perceptions around mixed emotions. While the phenomenon of mixed emotions continues to be a topic of research \cite{oh2022specificity}, we aim to replicate the original study with human participants in the future.

\textbf{Study 3 hypothesis validity}.
In this part we anticipated response similarity based on geographical proximity (that is assumed to lead to cultural similarity), and based on the literature supporting the likelihood of a similar behavior around mixed emotions in different types of societies. We emphasize that \textit{study 3} did not aim to prove the existence of this phenomenon, rather understand whether cultural similarities are found in English-European and East-Asian languages.

\textbf{Biases}. We adhered to the protocol outlined by Miyamoto et al. \cite{miyamoto2010culture}, including survey instructions where the content of the situations introduced potential mixed emotions, which may have biased the results. However, the findings indicate that despite this uniform introduction of bias across all experiments, the models' responses did not demonstrate a consistent pattern.

\section*{Acknowledgments}
This research was partially supported by PID2022-138721NB-I00 Grant from the Spanish Ministry of Science, Research National Agency, and FEDER (UE) funds. We thank Sophie Wang, Tuan Ann Dinh, Joy Jee and Yale Kim, Parfait Atchadé, and Michael Dabis for translating the surveys into Chinese, Vietnamese, Korean, French, and German, respectively.

\bibliographystyle{IEEEtran}
\bibliography{IEEEabrv,bibliography}

\end{document}